\documentclass[12pt]{article}
\usepackage[sort]{natbib}

\usepackage[utf8]{inputenc} 
\usepackage[T1]{fontenc}    
\usepackage[colorlinks = true,
            linkcolor = blue,
            urlcolor  = blue,
            citecolor = blue,
            anchorcolor = blue]{hyperref}
            
\usepackage{color}

\usepackage{url}            
\usepackage{booktabs}       
\usepackage{amsfonts}       
\usepackage{nicefrac}       
\usepackage{amsmath}
\usepackage{graphicx}
\usepackage{subcaption}
\usepackage{wrapfig}
\usepackage[font=footnotesize,labelfont=bf]{caption}
\usepackage{libertine}
\usepackage{cleveref}
\usepackage[dvipsnames,table,xcdraw]{xcolor}
\usepackage{fullpage}
\usepackage{authblk}
\usepackage{graphicx}
\usepackage{amssymb}

\renewcommand{\epsilon}{\varepsilon}
\renewcommand{\phi}{\varphi}

\DeclareMathOperator{\cossim}{sim}

\newcommand{\calD}{\mathcal{D}}
\newcommand{\calL}{\mathcal{L}}
\newcommand{\calB}{\mathcal{B}}

\newcommand{\bbR}{\mathbb{R}}

\newcommand{\feat}{\mathrm{F}}
\newcommand{\heat}{\mathrm{T}}

\newcommand{\br}{\mathbf{r}}
\newcommand{\bx}{\mathbf{x}}

\newcommand{\bg}{\mathbf{g}}

\newcommand{\by}{\mathbf{y}}

\newcommand{\bh}{\mathbf{h}}

\newcommand{\netw}{\mathbf{f}}

\newcommand{\softargmax}{\sigma_{\mathrm{arg}}}

\newcommand{\losscons}{\ensuremath{\calL_{\text{eqv}}}}
\newcommand{\lossdivers}{\ensuremath{\calL_{\text{div}}}}
\newcommand{\losssimclr}{\ensuremath{\calL_{\text{contrast}}}}

\newcommand{\acc}{\mathrm{Acc}}

\newcommand{\BK}[1]{ {\left( #1 \right)} }

\newcommand{\norm}[1]{\left\Vert #1 \right\Vert}

\begin{document}

\title{
Pretrained equivariant features improve unsupervised landmark discovery
}

\author[]{Rahul Rahaman\thanks{rahul.rahaman@u.nus.edu}}
\author[]{Atin Ghosh\thanks{atin.ghosh@u.nus.edu}}
\author[]{Alexandre H. Thiery\thanks{a.h.thiery@nus.edu.sg}}
\affil[]{Department of Statistics and Applied Probability, \\National University of Singapore}

\maketitle

\begin{abstract}
Locating semantically meaningful landmark points is a crucial component of a large number of computer vision pipelines.
Because of the small number of available datasets with ground truth landmark annotations, it is important to design robust unsupervised and semi-supervised methods for landmark detection.
Many of the recent unsupervised learning methods rely on the equivariance properties of landmarks to synthetic image deformations. Our work focuses on such widely used methods and sheds light on its core problem, its inability to produce equivariant intermediate convolutional features. This finding leads us to formulate a two-step unsupervised approach that overcomes this challenge by first learning powerful pixel-based features and then use the pre-trained features to learn a landmark detector by the traditional equivariance method. Our method produces state-of-the-art results in several challenging landmark detection datasets such as the BBC Pose dataset and the Cat-Head dataset. It performs comparably on a range of other benchmarks.

\end{abstract}


\section{Introduction}\label{sec:intro}
Producing models that learn with minimal human supervision is crucial for tasks that are otherwise expensive to solve with manual supervision. Discovering landmarks within natural images falls into this category since manually annotating landmark points at the pixel level scales poorly with the number of annotations and the size of the dataset. Recent advancements in Deep Learning have brought significant performance improvements to various computer vision tasks. This work focuses on using deep-learning models for unsupervised discovery of landmarks, as adopted in several recent articles \citep{factorized_embed, conditionalImageGen, vector_exchange, landmarkAsStructuralRepr}.\\

{\it Equivariance}, the fundamental concept most modern unsupervised landmark discovery approaches are based upon, formalizes the intuitive property that, when an image is deformed through the action of a geometric transformation, semantically meaningful landmarks associated with that image should also be deformed through the action of the same deformation.
Most recent approaches for unsupervised landmark discovery leverage this equivariance property of landmark points.\\

The tasks of discovering and detecting semantically meaningful landmarks have been extensively studied. Developed before the advent of more recent deep-learning techniques and among the large number of proposed methods for solving these problems, it is worth mentioning the supervised and weakly-supervised learning approaches based on templates \citep{face_pose_landmark_in_the_wild, deformation_model_field}, active appearance \citep{active_appearance, active_appearance_revisited, automatic_feature_localization}, as well as the regression-based models \citep{boosted_regression_graph, conditional_regression_forests, face_alignment_regression, face_alignment_3000_fps}.
Most of the recently proposed methodologies such as Cascade CNN \citep{cascade_cnn}, CFAN \citep{coarse_to_fine_CFAN}, coordinate regression \citep{coordinate_reg} and several variants \citep{landmark_deep_multi_task, face_alignment_auxiliary_attrib, deep_deformation_landmark_localization, recurrent_attentive_refinement, estimation_under_occlusion, recurrent_3d_2d} leverage deep-convolutional feature extractors for solving the task of supervised landmark detection. \\

In contrast, the approach described in this text is fully unsupervised and yet can perform more robustly than several supervised approaches.
Among other unsupervised landmark discovery pipelines, the approach of \citet{factorized_embed} was one of the first ones to leverage the equivariance property of landmarks. Equivariance was also used to discover 3D landmarks in the work of \citet{3d_keypoint_equivariance}. Several other approaches are based on the creation of representations that disentangle geometric structures from appearance. Such approaches can be seen in the work of \citet{geometric_matching}, Warpnet \citep{kanazawa2016warpnet}, as well as  \citet{disentangleAppearDeform}. In the work of  \citet{denseImageLabelling}, objects are mapped to labels for capturing structures in the image. Similar ideas can also be found \citep{vector_exchange}. \\

Several approaches attempt to disentangle the latent representation of images into visual and structural content by comparing representation between related images. These related images can be obtained through synthetic image deformations, or by exploiting the temporal continuity of video streams. \citet{conditionalImageGen} reconstructs images by combining the visual and the structural content to discover landmarks. Simultaneous image generation and landmark discovery can also be seen in \citep{landmarkAsStructuralRepr,reed_learning_where_to_draw, auto_regressive_density_estimation}. Although this set of methods has proven to be successful in some situations, the reliance on reconstructing high-resolution images is problematic in several aspects: these methods require high-capacity models, are often slow to train, and perform sub-optimally in situations where images are corrupted by a high level of noise or in the presence of artifact, as is, for example, common in medical applications. The proposed approach does not rely on image reconstruction and the numerical experiments presented in Section \ref{sec.experiments} demonstrate that it can still perform competitively on challenging tasks involving human-pose estimations, a task generally solved by leveraging the temporal continuity of adjacent video frames.\\

%
%
\noindent
{\bf Contributions.}
In this article, we take a closer look at the equivariance property of intermediate representations within deep learning models for landmark discovery. 
%
\begin{itemize}
\item 
We empirically demonstrate that standard landmark discovery approaches are not efficient at enforcing that the intermediate representations satisfy the equivariance property. To this end, we define a metric similar to \textit{cumulative error distribution curves} that we use to quantify the equivariance of convolutional features. 
\item 
We propose a two-step approach for landmark discovery. In the first step, powerful equivariant features are learned through a contrastive learning method. In a second step, these equivariant features are leveraged within more standard unsupervised landmark discovery pipelines. Our method is able to find semantically meaningful and consistent landmarks, even in challenging datasets with high inter-class variations.
\item 
Our numerical studies indicate that the proposed methodology outperforms previously proposed approaches in the difficult task of finding human body landmarks in the BBC Pose dataset, facial landmarks in the Cat-head dataset. Our method performs comparably to other unsupervised methods in the CelebA-MAFL facial landmark dataset.
\end{itemize}


\section{Background}\label{sec:background}
Landmarks can be defined as a finite set of points that define the structure of an object. A particular class of objects is expected to have the same set of landmarks across different instances.
In many scenarios, the variations in the locations of the landmarks across different instances of the same object can be used for downstream tasks (e.g., facial recognition, shape analysis, tracking).\\

%
%
Consider an image $\bx \in \bbR^{H\times W \times C}$ represented as a tensor with height $H$, width $W$ and number of channels $C$. This image can also be thought of as (the discretization of) a mapping from $\Lambda \triangleq [0, H] \times [0, W]$ to $\mathbb{R}^C$. We are interested in designing a landmark extraction function $\by(\bx) = [\by_1(\bx), \ldots, \by_K(\bx)] \in \Lambda^K$ that associates to the image $\bx$ a set of $K \geq 1$ landmarks $\{\by_k(\bx)\}_{k=1}^K$ where $\by_k(\bx) \in \Lambda$ for $1 \leq k \leq K$. In this work, the mapping $\by: \mathbb{R}^{H \times W \times C} \to \Lambda^K$ is parametrized by a neural network, although it is not a requirement. Occasionally, in order to avoid notational clutter, we omit the dependence of $\by$ on the input image $\bx$.\\


%
%
Consider a bijective mapping $\bg:\Lambda \to \Lambda$ that describes a geometric deformation within of the domain $\Lambda$. Examples include translations, rotations, elastic deformations, with a proper treatment of the boundary conditions. To any function $\phi: \Lambda \to \mathbb{R}^C$ can be associated the function $\bg^\sharp \circ \phi: \Lambda \to \mathbb{R}^C$ defined as $\bg^\sharp \circ \phi(\lambda) = \phi(\bg^{-1}(\lambda))$ for all $\lambda \in \Lambda$. Since images can be thought of as discretizations of functions from $\Lambda$ to $\mathbb{R}^C$, the function $\bg^\sharp$ also defines, with a slight abuse of notation, a deformation mapping on the set of images. With these notations, the function $\bg$ describes the transformation of coordinates (i.e. elements of $\Lambda$), while the function $\bg^\sharp$ describes the action of the transformation $\bg$ when applied to all pixels of a given image. A landmark extraction function $\by_k: \mathbb{R}^{H \times W \times C} \to \Lambda$ is said to satisfy the {\it equivariance} property if the condition
\begin{align} \label{eqn:consistency}
\by_k[\bg^\sharp(\bx)] = \bg[\by_k(\bx)]
\end{align}
holds for all images $\bx \in \mathbb{R}^{H \times W \times C}$. Equation \eqref{eqn:consistency} translate the intuitive property that, when an image is deformed under the action of a deformation mapping $g^\sharp$, the associated landmarks are deformed through the action of the same deformation.\\

In this article, we follow the standard approach \citep{conditionalImageGen, factorized_embed} of implementing the landmark extraction mapping $\by_k: \mathbb{R}^{H \times W \times C} \to \Lambda$ as a differentiable function by expressing it as the composition of a heatmap function $\bh_k: \mathbb{R}^{H \times W \times C} \to \mathbb{R}^{H \times W}$ with the standard {\it spatial soft-argmax} \citep{chapelle2010gradient} function $\softargmax: \mathbb{R}^{H \times W} \to \Lambda$,
\begin{align*}
\bx 
\underset{\bh_k}{\longrightarrow}
\; \bh_k(\bx) 
\underset{\softargmax}{\longrightarrow}
\by_k(\bx) 
\end{align*}
so that $\by_k(\bx) = \softargmax \circ \bh_k (\bx)\in \Lambda$.
In order to train a landmark extraction model, it is consequently a natural choice to try to minimize the \textit{equivariance loss} 
\begin{align}\label{eqn:loss_consist}
\losscons(\bx, \bg) = \frac{1}{K} \sum_{k=1}^K 
\norm{ \by_k[\bg^\sharp(\bx)] \; - \; \bg[\by_k(\bx)]  }^2.
\end{align}
This formulation was first used by \citet{factorized_embed}, and then subsequently adopted by several other articles \citep{vector_exchange, landmarkAsStructuralRepr}. Unfortunately, minimizing the equivariance loss alone generally leads to a trivial solution: all landmarks coincide. In practice, it is consequently necessary to consider an additional \textit{diversity loss}
\begin{align}\label{eqn:loss_diversity}
\lossdivers(\bx) = \sum_{\lambda \in \Lambda} \BK{\sum_{k \in K}[\sigma \circ \bh_k(\bx)]_{\lambda} - \max_{1 \leq k \leq K} [\sigma \circ \bh_k(\bx)]_{\lambda}}
\end{align}
where $\sigma: \mathbb{R}^{H \times W} \to (0,1)^{H \times W}$ is the usual {\it spatial softmax} function, although other choices are certainly possible. The notation $[\sigma \circ \bh_k(\bx)]_{\lambda}$ denotes the spatial softmax evaluated at the coordinate $\lambda \in \Lambda$. The loss \eqref{eqn:loss_diversity} was first proposed in \citet{factorized_embed} in conjunction with the equivariance loss to penalize the concentration of different heatmaps at the same coordinate. The diversity loss is often computed for small image patches rather than individual coordinates. Landmarks can be learned by directly minimizing the total loss $\lossdivers + \losscons$ on a dataset on unlabeled images and randomly generated deformation mappings $\bg$.

%
%
\section{Equivariance of intermediate representations}
\label{sec:analysis}
this section defines a measure of equivariance for convolutional features. This allows us to 
measure the equivariance of intermediate convolutional features learned by the set of methods described in Section \ref{sec:background}.\\

\noindent
\textbf{Equivariant features.} 
Fully Convolutional Neural Networks (FCNN) are leveraged in most modern deep learning based landmark discovery methods \citep{conditionalImageGen, factorized_embed, vector_exchange}. Let $\bx \in \bbR^{H\times W\times C}$ be an image and $\feat(\bx) = \netw \in \bbR^{H\times W\times D}$ be a convolutional feature obtained from $\bx$ by passing it through a fully convolutional network $\feat(\cdot)$. One can always assume, using a spatial change of resolution if necessary, that the convolutional features have the same dimensions as the input image $\bx$ and can thus be parametrized by $\Lambda \equiv [0,H] \times [0,W]$. For any 2D coordinate $\lambda \equiv (u, v) \in \Lambda$, the vector $\netw(\lambda) \in \bbR^D$ can be thought of as a $D$-dimensional representation of a small region (i.e. receptive field) of the input image $\bx$ centred at $\lambda \equiv (u,v) \in \Lambda$. This remark motivates a natural extension of the notion of equivariance for convolutional features. Consider two images $\bx$ and $\bx'$ related by a geometric deformation $\bg: \Lambda \to \Lambda$ in the sense that $\bx' = g^\sharp(\bx)$, as well as the associated convolutional features defined as $\netw = \feat(\bx)$ and $\netw' = \feat(\bx')$. The convolutional feature extractor would be perfectly equivariant if the extracted features $\netw$ and $\netw'$ were such that $\netw(\lambda) = \netw'(\bg[\lambda])$ for any location $\lambda \in \Lambda$. A landmark extraction function, as described in 
Section \ref{sec:background}, that exploits equivariant convolutional features would directly inherit the equivariance properties of these features.
Note, though, that training a landmark extractor through the loss function described in Equation \eqref{eqn:loss_consist} only attempts to enforce the equivariance property of the final landmarks. Consequently, it is natural to investigate whether {\bf (i)} this equivariance property naturally percolates through the convolutional neural network and leads to the creation of equivariant convolutional features {\bf (ii)} there are more efficient training methodologies for enforcing the equivariance property throughout the neural network model.\\

\noindent
\textbf{Measure of equivariance:}
Let $\netw, \netw'$ be the convolutional features associated to two images $\bx$ and $\bx' = \bg^\sharp(\bx)$ related through the deformation function $\bg:\Lambda \to \Lambda$. As described previously, one can always assume that the convolutional features and the images themselves have the same dimension and can, consequently, be parametrized by $\Lambda \equiv [0,H] \times [0,W]$. To quantify the equivariance of the convolutional features, it is computationally impractical to require the property $\netw(\lambda) = \netw(\bg[\lambda])$ to hold for all $\lambda \in \Lambda$. Furthermore, it is desirable to focus on the equivariance property of a few well chosen locations.

\begin{figure*}[ht]
\centering
\includegraphics[width=1.0\textwidth]{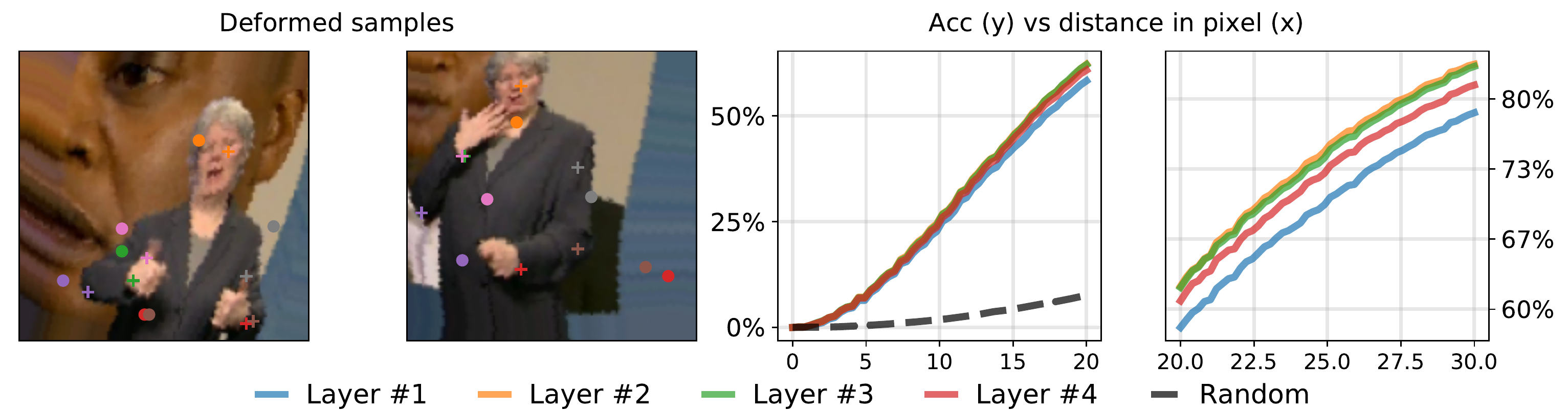}
\caption{Equivariance of intermediate convolutional features when trained by the \textit{end-to-end} approach. The first two images show deformed BBC Pose images with predicted (o) and ground truth (+) locations of $J=7$ annotated landmarks (head, two wrists, two elbows, two shoulders) found by maximizing similarity \eqref{eqn:find_max_match} from the features extracted from {\it Layer 1}, i.e. created by the feature extractor $\feat(\cdot)$. The plots on the right show the accuracy curves  \eqref{eqn:accuracy_distance} of several intermediate convolutional representations. The black dashed-line is a baseline consisting of random predictions. Even at a threshold distance $d=20$ pixels, the accuracy of most intermediate representations is as low as $\acc(d=20) \approx 57\%$. The rightmost \textit{zoomed-in} plot shows that the feature of {Layer 1}, i.e. the output of the feature extractor $\feat(\cdot)$, has the worst performance among the four considered representations.
}
\label{image:bbcpose_empirical_analysis}
\end{figure*}

To this end, in our study of the equivariance properties of intermediate convolutional features, we exploit a labelled dataset $\mathcal{D}$ whose images have been annotated with $J \geq 1$ semantically meaningful landmark points: to each image $\bx$ is associated $\by_1, \ldots, \by_J \in \Lambda$ landmarks. For such an image $\bx$ and a geometric deformation $\bg$, consider the deformed image $\bx' = \bg^\sharp(\bx)$ as well as the associated convolutional features $\netw$ and $\netw'$. For each landmark $\by_j$ associated to $\bx \in \mathcal{D}$, we consider the location $\widehat{\by}'_j \in \Lambda$ in the deformed image $\bx'$ whose representation is the most similar to $\netw'(\by_j)$,
\begin{align}
\label{eqn:find_max_match}
\widehat{\by}'_j = \underset{\lambda \in \Lambda}{\mathbf{argmax}} \; \cossim\big( \netw'(\lambda), \, \netw(\by_j) \big).
\end{align}
In \eqref{eqn:find_max_match} we have used the cosine similarity,
\begin{align}
\label{eq.def.sim}
\cossim(\netw_1, \netw_2) \triangleq
\biggl< \frac{\netw_1}{\|\netw_1\|}, \, \frac{\netw_2}{\|\netw_2\|} \biggl>,
\end{align}
although other choices are indeed possible. For discriminative and equivariant convolutional features, we expect the Euclidean distance between $\widehat{\by}'_j$ and $\by_j$
to be small. Consequently, for a distance threshold $d > 0$, the accuracy at threshold $d$ is defined as
\begin{align}\label{eqn:accuracy_distance}
\acc(d) 
\triangleq 
\frac{1}{|\calD| \cdot |J|} \sum_{j \in J,\bx_i \in \calD} \mathbf{I}\big( \| \widehat{\by}'_{i,j} - \by'_{i,j} \| \le d \big).
\end{align}
where $\widehat{\by}'_{i,j}$ and  $\by'_{i,j}$ denote the $j$-th landmark quantities associated to image $\bx_i \in \mathcal{D}$.
In the sequel, the curve $\{(d, \acc(d)): d\!\geq\!0\}$ will be referred to as the \textit{accuracy curve}. An example is depicted in Figure \ref{image:bbcpose_empirical_analysis}.\\

\noindent
\textbf{Features learned from landmark equivariance.}
The accuracy metric \eqref{eqn:accuracy_distance} can be used to investigate whether the intermediate convolutional features learned by directly minimizing the equivariance loss \eqref{eqn:loss_consist} combined to the diversity regularization \eqref{eqn:loss_diversity}, i.e. $\losscons+\lossdivers$, also exhibit good equivariance properties. 
The approach consisting in directly minimizing the combined loss $\losscons+\lossdivers$ is referred to as the \textit{end-to-end} learning approach, in contrast to our proposed two-step approach, described in Section \ref{sec:method}.
Furthermore, we use standard so-called {\it hourglass} convolutional neural architectures as recommended by most recent work on landmark discovery \citep{conditionalImageGen, vector_exchange, landmarkAsStructuralRepr}.
Such a model architecture $\heat \circ \feat$ can be described as the composition of a convolutional feature extractor $\feat: \mathbb{R}^{H \times W \times C} \to \mathbb{R}^{H \times W \times D}$ that produces $D$-dimensional spatial features followed by another convolutional network $\heat: \mathbb{R}^{H \times W \times D} \to \mathbb{R}^{H \times W \times K}$ that transforms these features into $K$ heatmaps $(\bh_1, \ldots, \bh_K)$ where $\bh_k \in \mathbb{R}^{H \times W}$. As described in Section \ref{sec:background}, these $K$ heatmaps are finally transformed into $K$ landmark points $\softargmax(\bh_1), \ldots, \softargmax(\bh_K) \in \Lambda$ through a {\it spatial soft-argmax} operation.
For our experiments, we chose $K = 50$ landmark points and convolutional features of dimension $D = 64$. 
We trained in an \textit{end-to-end} manner the hourglass network $\heat \circ \feat$ on the BBC Pose dataset \citep{bbcpose} training dataset for landmark discovery.
The accuracy metric \eqref{eqn:accuracy_distance} was then computed on the BBC Pose dataset's test set (i.e. not used during training). To this end, we chose a set of $J=7$ manually annotated human body landmarks: head, two shoulders, two elbows, and two wrists. The images were resized to a resolution $128 \times 128$ pixels.

Figure \ref{image:bbcpose_empirical_analysis} presents examples from the BBC Pose test dataset where random rotations and random elastic deformations were applied. The circles denote the locations $\widehat{\by}'_i$ estimated through Equation \eqref{eqn:find_max_match} and the crosses indicate the ground truth locations $\bar{\by}'_i$; these locations were obtained by training a landmark extractor in an {\it end-to-end} manner.
The right-most plots of Figure \ref{image:bbcpose_empirical_analysis} report the accuracy curves $\{(d, \acc(d)) \, : \, d > 0)\}$ estimated from a set of $J=7$ annotated landmarks. Each of the four accuracy curves is associated to one of four intermediate convolutional features coined {\it Layer 1} to {\it Layer 4}. Here, {\it Layer 1} is the output of the feature extractor $\feat(\cdot)$ while {\it Layer (2, 3, 4)} are representation situated in between {\it Layer 1} and the final heatmaps generated through the complete network $\heat \circ \feat$.
For comparison, the black dashed line shows the performance associated with locations $\widehat{\by}'_i$ generated uniformly at random within $\Lambda=[0,H] \times [0,W]$.
The results indicate that, even though the intermediate convolutional features do possess some degree of equivariance (i.e. better than random), the performance is relatively poor with an accuracy $\acc(d) = 57\%$ for a distance threshold of $d=20$ pixels.

%
%
\section{Method}
\label{sec:method}
The previous section demonstrates that the direct minimization of the combined loss $\losscons+\lossdivers$ is not efficient at inducing equivariance properties to the intermediate convolutional features. In contrast, the first step of the methodology described in this section aims at directly enforcing that the intermediate features enjoy enhanced equivariance properties. In a second step, these pre-trained features are leveraged within more standard landmark discovery pipelines. 
Although we can choose to pre-train any of the convolutional features of the complete landmark extractor $\heat \circ \feat$, our experiments indicate that it is most efficient to concentrate on the final convolutional features (i.e. {\it layer 1} with the terminology of the previous section) generated by the feature extractor network $\feat(\cdot)$. As depicted in the right-most plot of Figure \ref{image:bbcpose_empirical_analysis}, when {\it end-to-end} training is used, the {\it layer 1} is the worst in terms of equivariance among the four layers considered in the previous section. At a heuristic level, this can be explained by the fact that the {\it layer 1} is the further ``away" from the training signal among the four layers considered.

Our proposed methodology proceeds in two steps.
{\bf Step 1:} train the feature extractor convolutional network $\feat$ with the contrastive learning approach described in the remainder of this section. 
{\bf Step 2:} freeze the parameters of the feature extractor network $\feat(\cdot)$ and train the complete network $\heat \circ \feat(\cdot)$ with a standard landmark discovery approach, i.e. minimize the combined objective $\losscons + \lossdivers$.\\

\noindent
\textbf{Step 1: training the feature extractor $\feat(\cdot)$.} 
We employ a contrastive learning framework similar to simCLR \citep{simCLR} to train the features extractor. To this end, we leverage {\bf (1)} random geometric deformations $\bg: \Lambda \to \Lambda$ that only change the locations of the pixels (eg. rotations, dilatations, elastic deformations) {\bf (2)} random appearance changes denoted by $\br: \mathbb{R}^{H,W,D} \to \mathbb{R}^{H,W,D}$ that do not change the locations of the pixels (eg. noise addition, intensity change, contrast perturbation).
For a batch of images $(\bx_1, \ldots, \bx_B) \equiv \mathcal{B}$, we consider augmented versions defined as $\bx'_b = (\br_b \circ \bg^\sharp_b)(\bx_b)$ augmented by random generated geometric and appearance augmentations $\bg_b$ and $\br_b$. Furthermore, for each image $\bx_b \in \mathcal{B}$ in the batch, we also consider $K \geq 1$ randomly selected spatial locations $\lambda_{b,1}, \ldots, \lambda_{b,K} \in \Lambda$: the location $\lambda_{b,k}$ is transform to $\lambda'_{b,k} \equiv \bg_b(\lambda_{b,k})$ under the action of the geometric deformation $\bg_b$. Finally, these pairs of images are passed through the feature extractor $\feat: \mathbb{R}^{H \times W \times C} \to \mathbb{R}^{H \times W \times D}$ and $D$-dimensional features are extracted at the locations $\lambda_{b,k}$ and $\lambda'_{b,k}$ for $1 \leq b \leq |\mathcal{B}|$ and $1 \leq k \leq K$. We have 
$\netw_{b,k} \triangleq \feat(\bx_b)[\lambda_{b,k}] \in \mathbb{R}^D$ and
$\netw'_{b,k} \triangleq \feat(\bx'_b)[\lambda'_{b,k}] \in \mathbb{R}^D$.
The feature extractor $\feat$ is trained by minimizing the 
contrastive loss $\losssimclr$ defined as
%
%
\begin{align}
\label{eqn:contrastive_loss}
\calL_{\text{contrast}} \triangleq -\frac{1}{|\calB| \cdot K} \sum_{b,k} \log p_{b,k} + \log p'_{b,k} 
\end{align}
where the quantities $p_{b,k}$ and $p'_{b,k}$ are defined as
\begin{align*}
\left\{
\begin{aligned}
p_{b,k} &\triangleq \frac{s_{\tau}(\netw_{b,k}, \netw'_{b,k})}{ \sum\limits_{(n,i) \neq (b,k)} s_{\tau}(\netw_{b,k}, \netw_{n,i}) + \sum\limits_{n,i} s_{\tau}(\netw_{b,k}, \netw'_{n,i})}\\
p'_{b,k} &\triangleq \frac{s_{\tau}(\netw_{b,k}, \netw'_{b,k})}{ \sum\limits_{(n,i) \neq (b,k)} s_{\tau}(\netw'_{b,k}, \netw'_{n,i}) + \sum\limits_{n,i} s_{\tau}(\netw'_{b,k}, \netw_{n,i})}
\end{aligned}
\right.
\end{align*}
and $s_{\tau}(\mathbf{a}, \mathbf{b}) \triangleq \exp\{\cossim(\mathbf{a}, \mathbf{b})/\tau\}$ is the exponential similarity between $\mathbf{a}, \mathbf{b} \in \bbR^D$ scaled by temperature $\tau$. The similarity $\cossim(\cdot, \cdot)$ is defined in Equation \eqref{eq.def.sim}.
Minimizing the contrastive loss not only ensures equivariance, it also encourages dissimilarity between features belonging to different parts of the image. The landmark detector trained in the second step described below benefits from both the consistency and contrast of these features.\\

\noindent
\textbf{Step 2: training the landmark detector.} 
After training the feature extractor $\feat(\cdot)$ as described in the previous section, its parameters are frozen. The entire network $\heat \circ \feat$ is then unsupervisedly trained in a standard fashion for landmark discovery. In this text, we follow the approach of \citet{factorized_embed} although other approaches are possibles.
We use a fully convolutional architecture with residual connections and minimize the combined loss $\losscons + \lossdivers$.
Furthermore, to encourage the heatmaps to be more concentrated, we also penalize the variance of the heatmaps. In other words, we consider an additional regularization term expressed as
\begin{gather*}
\calL_{\text{variance}} \triangleq 
\frac{1}{|\mathcal{B}|}\sum_{b=1}^{|\mathcal{B}|} \; 
\mathbf{Tr} \BK{ \mathbf{Cov} \left[ \sigma \circ \heat \circ \feat(\bx_b) \right] }
\end{gather*}
In the above Equation, $\sigma(\cdot)$ denotes the {\it spatial softmax} operation that transforms a heatmap $\heat \circ \feat(\bx_b)$ on $\Lambda$ into a probability measure on $\Lambda$. The notation $\mathbf{Cov}$ designates the covariance matrix. We defer additional details of implementation to the supplementary.\\

\begin{figure*}[ht]
\centering
\includegraphics[width=1.0\textwidth]{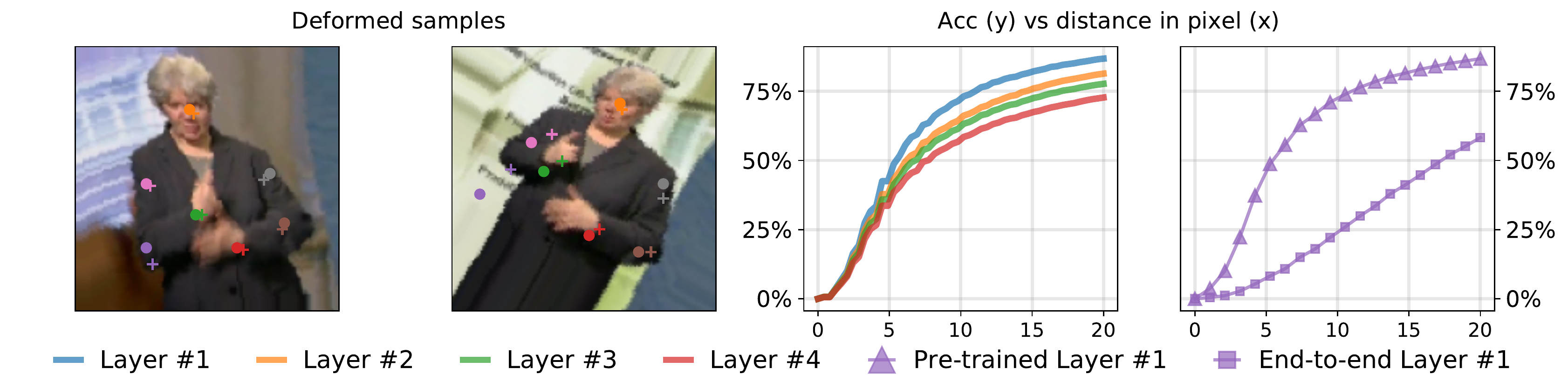}
\caption{Performance gains offered by the two-steps approach. The first two images show deformed BBC Pose images with predicted (o) and ground truth (+) locations of $J=7$ annotated landmarks (head, two wrists, two elbows, two shoulders) found by maximizing similarity \eqref{eqn:find_max_match} from the features extracted from {\it Layer 1}, i.e. created by the feature extractor $\feat(\cdot)$.
In comparison to Figure \ref{image:bbcpose_empirical_analysis}, predictions are significantly more accurate.
The third plot shows the accuracy curves \eqref{eqn:accuracy_distance} of several intermediate convolutional representations.
The right-most plot compares the equivariance properties of {\it Layer 1} when trained with an {\it end-to-end} method and with the proposed two-step approach.}
\label{image:bbcpose_empirical_analysis_pretrained}
\end{figure*}

\subsection{Implementation Details}
We choose the network $\feat$ to be an hourglass encoder-decoder network. The encoder contains four convolution-maxpool blocks, each of which halves the spatial dimension of its input. Symmetrically, the decoder block expands the spatial dimension to produce final features of the same spatial dimension as that of the original input image. We use skip-connections to join the encoder and decoder. Each of the convolution blocks contains two convolution units separated by non-linear activation ReLU and one max-pool/up-sample unit. The final channel dimension of the features $\feat(\bx)$ is $D=64$. The landmark network $\heat$ is fully convolutional with the same intermediate spatial dimension as that of the image. For contrastive learning, we used a temperature of $\tau = 0.1$.\\

We first train the network $\feat$ with the contrastive loss described in equation 6 of the main draft on all the datasets. The image deformation set includes random affine transformations, random crop and zoom operations, image augmentations related to intensity and contrast, as well as elastic deformations. Next, we freeze the parameters of the network $\feat$ and train the landmark detector. For this, we use the combined loss $\losscons + \lossdivers + \calL_{\text{variance}}$ with the image deformation as random rotation and elastic deformation. Note that this combined loss was also used to train the end-to-end network presented in ablation.\\

For the diversity loss, we choose image patches of size $8\times 8$ for images of size $128\times 128$ and $10\times 10$ patches for $100\times 100$ images.
We use the Adam optimizer with a learning rate of $3.10^{-4}$ and LR decay of $0.9$ every epoch for both training steps. For both steps, we train our model for 20 epochs. We also use $L^2$ regularization of strength $5.10^{-4}$.
To check the performance of unsupervisedly discovered landmarks, we use a linear regressor without bias to predict the ground truth annotations. The regressor is always trained on the training set and then evaluated on the test dataset. We keep the training and test split the same as previous works. We use Ridge regression as our choice of regressor with regularization $\alpha=0.1$.

\section{Experiments}
\label{sec.experiments}
%
In this section, we illustrate the performance gains in terms of equivariance of the intermediate features when adopting our proposed two-step method. Figure \ref{image:bbcpose_empirical_analysis_pretrained} shows some examples of deformed images from the BBC Pose dataset with the ground truth and predicted locations of $J=7$ landmarks: the setting is similar to the one of Figure \ref{image:bbcpose_empirical_analysis}. The predicted locations were estimated through maximizing the similarity, as described in Equation \eqref{eqn:find_max_match}.
The third plot shows the accuracy of the four intermediate representations {\it layers (1,2,3,4)} previously described at the end of Section \ref{sec:method}: note that all layers benefited in terms of equivariance from the two-step approach, even though only {\it layer 1} was pre-trained with contrastive learning.
The right-most plot compares the equivariance accuracy curve associated to {\it Layer 1} when the model is trained in an {\it end-to-end} fashion and when trained with our proposed approach: there is a considerable boost in equivariance accuracy \eqref{eqn:accuracy_distance}. At a distance threshold of $d=20$ pixels, the \textit{end-to-end} method leads to an accuracy of $\acc(d=20)=57\%$ while the proposed two-step approach leads to an equivariance accuracy of $\acc(d=20)=87\%$.

\subsection{Human body landmarks.}
Learning human body landmarks in an unsupervised manner poses some unique challenges. Unlike rigid and semi-rigid structures such as the human face, vehicles, or other non-deformable objects, the human body allows flexibility around different body joints. The BBC Pose dataset is comprised of $20$ videos of sign language interpreters. The positional variability of the different upper body parts during sign language communication leads to unique challenges. Following the setup of \citet{conditionalImageGen}, loose crops around the interpreters were resized to $128\times 128$ and used for our experiments.

\begin{figure*}[ht]
\centering
\includegraphics[width=0.9\textwidth]{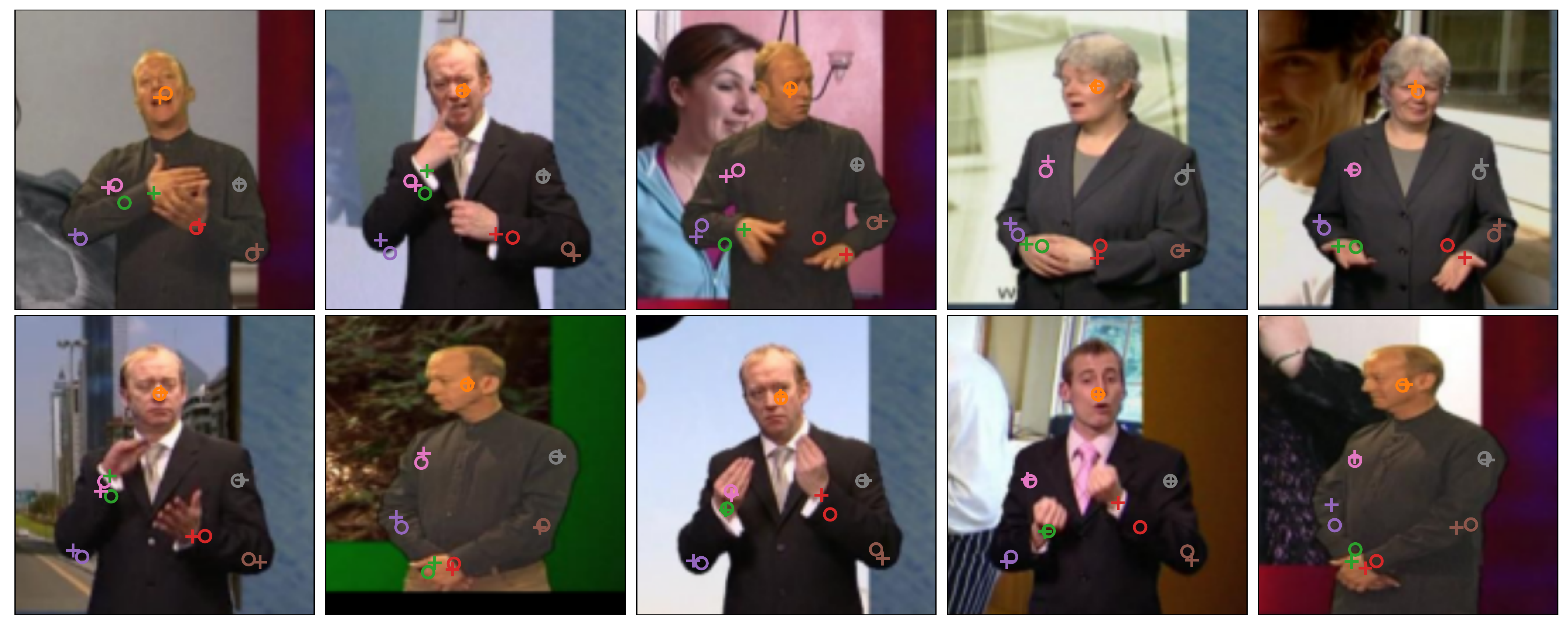}
\caption{Landmark prediction in BBC Pose dataset: Predicted (hollow circle: `$\circ$') vs. ground truth (`+') locations of the seven annotated landmarks (head, two shoulders, two wrists, two elbows) in BBC Pose dataset. Fifty unsupervised landmarks were used for this prediction. Visually our predictions and ground truth positions are quite close and are in fact coinciding in many examples.}
\label{image:bbcpose_predicted_vs_gt_final}
\end{figure*}

\noindent
{\bf Qualitative results.} We used the training part (ten videos) of the BBC Pose dataset to learn in an unsupervised manner $K=50$ landmarks with our proposed two-step method. After training, the discovered landmarks were computed on the validation dataset (five videos) and were used as features to train a linear regressor (without bias) to predict the actual location of the seven annotated landmarks. This regression model is then evaluated on the test dataset (five videos). Figure \ref{image:bbcpose_predicted_vs_gt_final} shows some example of the test dataset with predicted (o) and ground truth (+) locations of the landmarks.

\begin{table}[h]
    \centering
    \begin{tabular}{|c|c c c c c|}
    \hline
          & \multicolumn{5}{c|}{Proportion (\%) within $6$ pixels }\\
    \hline
        \textbf{Supervised} & Head & Wrst & Elbw & Shldr & Avg.\\
    \cline{1-1}
        \citet{Pfister_flowing_convnets} & $98.0$ & $\mathbf{88.4}$ & $77.1$ & $\mathbf{93.5}$ & $\mathbf{88.0}$\\
        \citet{Charles_domain_adaptation} & $95.4$ & $72.9$ & $68.7$ & $90.3$ & $79.9$\\
        \citet{chen_articulated_pose_graphical_model} & $65.9$ & $47.9$ & $66.5$ & $76.8$ & $64.1$\\
        \citet{Pfister_pose_estimation_gesture} & $74.9$ & $53.0$ & $46.0$ & $71.4$ & $59.4$\\
        \citet{yang_articulated_mixture_of_parts} & $63.4$ & $53.7$ & $49.2$ & $46.1$ & $51.6$\\
    \hline
        \textbf{Unsupervised} & & & & &\\
    \cline{1-1}
        \citet{conditionalImageGen} & $76.1$ & $56.5$ & $70.7$ & $74.3$ & $68.4$\\
        Ours & $\mathbf{99.4}$ & $33.5$ & $\mathbf{78.3}$ & $\mathbf{93.5}$ & $72.9$\\
    \hline
    \end{tabular}
\caption{Proportion of landmarks predicted within a Euclidean distance of $6$ pixels from the ground truth on image of size $128\times 128$ of the BBC-Pose dataset.
Our method outperforms all the supervised methods in detecting head, elbow and shoulder. Our model does not perform well on locating `wrists' mainly because unlike our method, the other methods including the unsupervised method leverages temporal continuity of video frames. }
    \label{tab:bbcpose_numerics}
\end{table}

\begin{table}[h]
    \centering
    \begin{tabular}{|c|c c|}
        \hline
            & \multicolumn{2}{c|}{Overall accuracy \%}\\
        \cline{2-3}
            $\#$ Training Samples & \citet{conditionalImageGen} & Ours\\
        \hline
            $5$ & - & $49.7 \pm 5.0$\\
            $10$ & - & $57.6 \pm 3.8$\\
            $50$ & - & $67.2 \pm 0.8$\\
            $100$ & - & $69.4 \pm 0.6$\\
            $500$ & - & $72.1 \pm 0.5$\\
            $1000$ (full dataset) & $68.4$ & $72.9$\\
        \hline
    \end{tabular}
\caption{Proportion of landmarks predicted within a Euclidean distance of $6$ pixels from the ground truth on image of size $128\times 128$ of the BBC-Pose dataset.
The table reports the perfmance as a function of the number of training samples used during training. Our proposed method outperforms previous unsupervised approaches with only $100$ samples.}
    \label{tab:bbcpose_sampleeff}
\end{table}

\noindent
{\bf Quantitative results.}
In table \ref{tab:bbcpose_numerics} we present our numerical results on the unsupervised landmark detection. The first few methods are fully supervised, and most of these methods leverage temporal continuity to track landmarks reliably across video frames. In contrast, \citet{conditionalImageGen} and ours are unsupervised and only use the ground truth annotations of the validation dataset to predict the locations of the annotated landmarks on the test dataset.
Following the evaluation procedure of \citet{conditionalImageGen}, we report the proportion of landmarks predicted within a Euclidian distance of $6$ pixels from the ground truth in an image of size $128\times 128$.\\

Table \ref{tab:bbcpose_numerics} shows the prediction score for different classes of upper-body joints, as well as the overall score. When averaged all the landmarks, our method outperforms three of the supervised methods and outperforms the unsupervised method of \citet{conditionalImageGen}. Furthermore, except for the \textit{wrists}, our method outperforms all the \textit{supervised} methods for predicting individual landmarks. The relatively poor performance for detecting the location of the \textit{wrists} stems from the fact that our method uses elastic deformations of the same image frame to learn equivariance. In contrast, most of the supervised methods exploit implicit temporal continuity while the method of \citet{conditionalImageGen} uses a different frame from the same video to obtain original and deformed image pair. Our methodology could certainly leverage recent image key-point matching methods in order to leverage more efficiently video modalities. Finally, table \ref{tab:bbcpose_sampleeff} reports the performance of our method as a function of the number of supervised samples from the validation dataset (total 1000 samples) used to fit the regression model. We can see that with just 100 supervised annotations, our method outperforms the approach of \citet{conditionalImageGen}.

\subsection{Facial landmarks}
For the task of learning facial landmarks, we implement our methodologies on the  \textit{CelebA} \citep{celebA} dataset (200k human face images of 10k celebrity) as well as the \textit{Cat Head} \citep{CatHeadDataset} dataset (9k images of cat faces). The celebA dataset contains $5$ annotated facial landmarks (two eyes, nose, two corners of mouth) whereas the cat dataset has $9$ annotated landmark points (six points in ears, two eyes, one mouth). The cat head dataset features more holistic variations than human face images and contains a considerable amount of occlusions. Similar to human body landmarks, we train our two-step landmark detection network on a training set, fit a linear regressor on a validation set and finally evaluate the performance of the regressor on a test dataset. In all our experiments related to facial landmarks, we adopt the popular metric \textit{inter-ocular distance normalized MSE},
\begin{gather}\label{eq:iod-mse}
\text{iod-mse} \; \triangleq \; 
\frac{1}{|\calD|\cdot|I|} \sum_{i \in I,\bx \in \calD} \frac{ \norm{\widehat{\by}_i - \bar{\by}_i}}{\text{Iod}(\bx)},
\end{gather}
where $\calD$ denotes the test dataset, $I$ is the set of manually annotated landmarks, $\widehat{\by}_i $ and $\bar{\by}_i$ are respectively the predicted and ground truth location of the $i^{\text{th}}$ landmark in image $\bx$, and $\text{Iod}(\bx)$ denotes the distance between the ground truth eyes in an image $\bx$.

\begin{figure*}[ht]
\centering
\includegraphics[width=1.\textwidth]{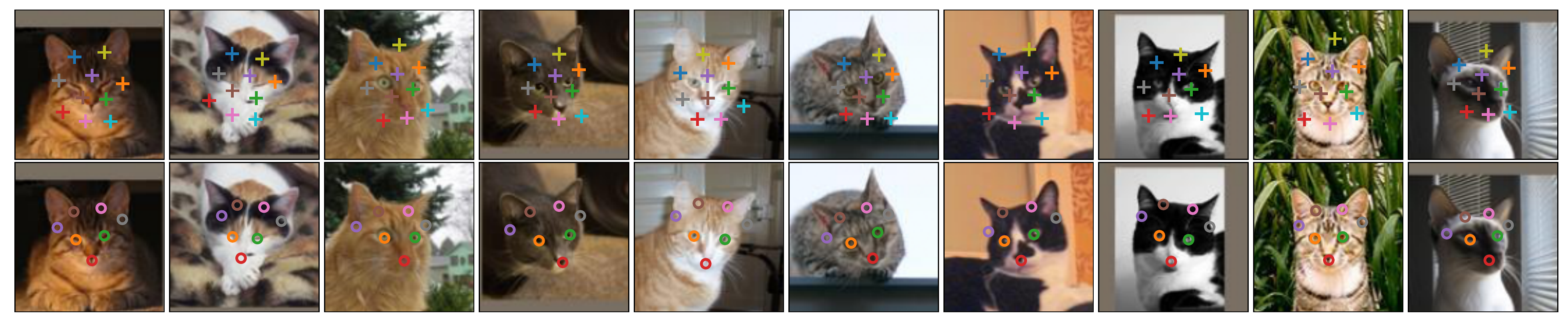}
\caption{Cat head dataset: the top row shows $K=10$ unsupervised landmarks points (+) predicted by our model. The bottom row shows seven facial landmark points (four in ears, two eyes, mouth) predicted by the ten unsupervised landmarks through a linear regressor. Visibly, this dataset offers a lot of inter-example variations due to different cat breeds and colors. Our unsupervised landmarks are semantically meaningful, and our supervised predictions are visibly accurate.}
\label{image:cat_head_examples}
\end{figure*}

\begin{table}[h]
    \centering
    \begin{tabular}{|c|c c|}
        \multicolumn{3}{c}{Performance on Cat head dataset}\\
    \hline
        Landmark count & $K=10$ & $K=20$\\
    \hline
        \citet{factorized_embed} & $26.76$ & $26.94$\\
        \citet{landmarkAsStructuralRepr} & $15.35$ & $14.84$\\
    \hline
        Ours ($\#$ Training Samples) & & \\
        $5$ & $21.5 \pm 2.4$ & $21.1 \pm 1.6$ \\
        $10$ & $18.9 \pm 0.6$ & $18.8 \pm 1.5$ \\
        $100$ & $16.2 \pm 0.3$ & $15.5 \pm 0.3$ \\
        $500$ & $15.2 \pm 0.2$ & $14.6 \pm 0.3$ \\
        $1000$ & $14.9 \pm 0.0$ & $14.3 \pm 0.1$ \\
        $7747$ (full dataset) & $\mathbf{14.59}$ & $\mathbf{13.80}$\\
    \hline
    \end{tabular}
    \caption{Prediction performance (inter-ocular distance normalized MSE expressed in $\%$) and sample efficiency of our method on Cat head dataset when $K=10$ and $K=20$ were unsupervisedly learned. Our method outperforms previous unsupervised models with only $500$ samples. The total number of training samples is more than 7000.}
    \label{tab:cat_head_numerical}
\end{table}

\noindent
\textbf{Cat head dataset.} We keep the size of the cat images to their original $100\times 100$ resolution and maintain the same train/test split as \citet{landmarkAsStructuralRepr}. Furthermore, we discard the two annotations corresponding to the end of the ears \citep{landmarkAsStructuralRepr,factorized_embed}. 

\textit{Qualitative results.} Figure \ref{image:cat_head_examples} shows some examples of cat head test dataset. The top row shows images annotated with the ten unsupervised landmark locations discovered by our model. The bottom row shows the result of predicting ground truth landmark locations of seven facial landmark points. It is worth noting that the inter-example variation in this dataset is more than any human face dataset as the images vary in \textit{pose}, \textit{color}, \textit{cat breed}, \textit{occlusion}.

\begin{table}[h]
    \centering
    \begin{tabular}{|c|c c c|}
            \multicolumn{4}{c}{Performance on MAFL dataset}\\
        \hline
             Landmark count & NA & $K=30$ & $K=50$\\
        \hline
            \textbf{Supervised} & & & \\
            CFAN \citep{coarse_to_fine_CFAN} & $15.84$ & - & -\\
            Cascaded CNN \citep{cascade_cnn} & $9.73$ & - & -\\
            TCDCN \citep{face_alignment_auxiliary_attrib} & $7.95$ & - & -\\
            MTCNN \citep{landmark_deep_multi_task} & $6.90$ & - & -\\
        \hline
            \textbf{Un/Self-supervised} & - & - & -\\
            \citet{factorized_embed} & - & $7.15$ & $6.67$\\
            \citet{denseImageLabelling} & $5.83$ & - & -\\
            \citet{disentangleAppearDeform} & $5.45$ & - & -\\
            \citet{sanchez} & $3.99$ & - & - \\
            \citet{landmarkAsStructuralRepr} & - & $3.16$ & -\\
            \citet{vector_exchange} & $2.86$ & - & -\\
            \citet{conditionalImageGen} & - & $\mathbf{2.58}$ & $\mathbf{2.54}$\\
        \hline
            Ours & - & $4.59$ & $4.31$\\
        \hline
    \end{tabular}
    \caption{Inter-ocular distance normalized MSE (expressed in $\%$) dataset as defined in Equation \eqref{eq:iod-mse}.}
    \label{tab:MAFL_numerical}
\end{table}
\begin{table}
   \centering
   \begin{tabular}{|c|c c|}
        \hline
            No. sample & \citet{conditionalImageGen} & Ours\\
        \hline
            1 & $12.89 \pm 3.2$ & $11.12 \pm 2.1$\\
            5 & $8.16 \pm 0.9$ & $8.82 \pm 0.4$\\
            10 & $7.19 \pm 0.4$ & $8.27 \pm 0.3$\\
            full data & $2.58$ & $4.59$\\
        \hline
    \end{tabular}
    \caption{Sample efficiency on MAFL. When the sample size is extremely low, our model performs comparably. The full training dataset contains 19000 images of human face.}
    \label{tab:MAFL_sampleeff}
\end{table}

\textit{Quantitative results.} Table \ref{tab:cat_head_numerical} shows performance of our method. The performances of two unsupervised methods are listed at the top of the table. The bottom part of the table reports the performance of our proposed approach as a function of the number of supervised samples. We report both the mean and standard deviation in order to account for the variability in the regression training set. Our method can outperform previous unsupervised methods when only $500$ supervised samples out of the total available $7,747$ training samples are utilized.\\

\noindent
\textbf{CelebA dataset.} The CelebA images were resized to $128\times 128$ as done by most works \citep{conditionalImageGen}. The \textit{MAFL} dataset, a subset of the CelebA dataset, has a training set of 19k images and a test set of 1k images. Like previous works, we take our unsupervised model trained on CelebA and then train a linear regressor to predict the MAFL training set's manual annotations.

\begin{figure*}[ht]
\centering
\includegraphics[width=1.\textwidth]{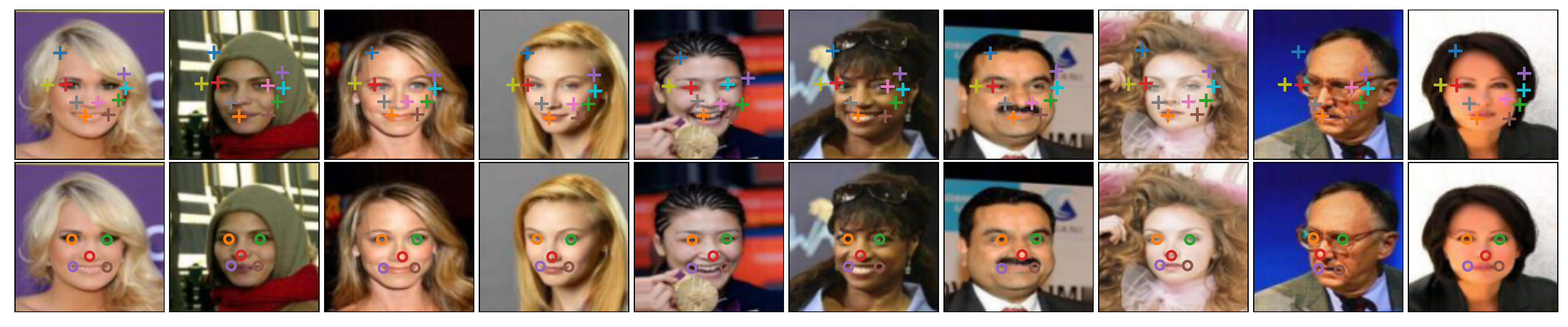}
\caption{MAFL facial landmarks: Examples from MAFL test dataset. The top row shows ten unsupervised landmarks points (+) predicted by our model. The bottom row shows five predicted facial landmark points (eyes, nose, mouth) learned from the 10 unsupervised landmarks.}
\label{image:mafl_examples}
\end{figure*}

\textit{Qualitative results.} Some predictions generated by our approach are depicted in Figure \ref{image:mafl_examples}. The top row shows ten unsupervised landmarks learned for each image. The bottom row shows five landmarks as predicted by the unsupervised landmarks through a linear regressor without bias.

\textit{Quantitative results.} Table \ref{tab:MAFL_numerical} reports the performance of several supervised, unsupervised, and semi-supervised methods. 
Although our performance is not better than the previous state-of-the-art in terms of iod-mse\%, for an image size of $128\times 128$, the difference in actual MSE in terms of pixel distance is marginal. Table \ref{tab:MAFL_sampleeff} presents the performance of the proposed approach as a function of the number of supervised samples utilized to train the regressor: our model performs comparably with the \textit{SOTA}.

\begin{table}[ht]
    \centering
    \small
    \begin{tabular}{|c | c c c c c| c c | c c |}
        \cline{2-10}
            \multicolumn{1}{c}{ } &
            \multicolumn{5}{|c}{BBC Pose Accuracy (\%)} &
            \multicolumn{2}{c}{MAFL mse (\%)} &
            \multicolumn{2}{c|}{Cat head mse (\%)}\\
        \hline
              Method & Head & Wrst & Elbw & Shldr & Avg.
              & 30 & 50 & 10 & 20\\
        \hline
            \citet{factorized_embed} & -- & -- & -- & -- & -- 
                                                   & $7.15$ & $6.67$ & $26.76$ & $26.94$\\
            End-to-end & $72.0$ & $15.5$ & $45.5$ & $57.0$ & $44.0$
                       & $6.57$ & $5.97$ & $22.97$ & $23.42$ \\
            Pre-training & $\mathbf{99.4}$ & $\mathbf{33.5}$ & $\mathbf{78.3}$ & $\mathbf{93.5}$ & $\mathbf{72.9}$
                                & $\mathbf{4.59}$ & $\mathbf{4.31}$ & $\mathbf{14.74}$ & $\mathbf{13.80}$\\
        \hline
    \end{tabular}
    \caption{Result of our ablation study. In the first row we show the performance of the original work by \citet{factorized_embed}. In the second row we adopt the end-to-end training method of equivariance training but replace the architecture with our own model $(S:T)$. The third row shows the result of our proposed two-stage method. The architecture and the losses used in second and third row is same. Hence, the improvement from first to second row is due to the architecture and the final improvement from second to third row is due to our proposed pre-training method.}
    \label{tab:ablation}
\end{table}

\begin{figure}[ht]
\centering
\includegraphics[width=1.0\textwidth]{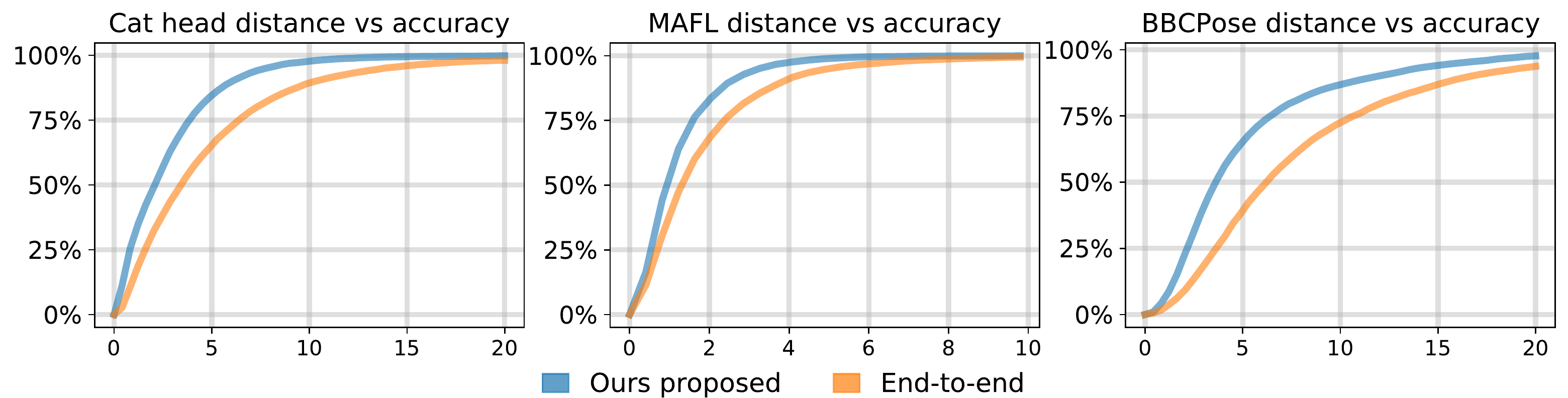}
\caption{Accuracy curve of final predicted landmarks that are obtained from linear regressor fitted on top of unsupervised landmarks. We compare the performance of our proposed pre-training vs. end-to-end training method. Similar to previous accuracy curves, the x-axis denotes the distance $d$ and y-axis denotes the corresponding accuracy $\acc(d)$. }
\label{fig:final_cummulative_error}
\end{figure}

\subsection{Ablation study}
We perform our ablation table \ref{tab:ablation} that lists three methods.
We compare our proposed method to the one of \citet{factorized_embed} since it is the most similar.
The second row reports the performance of the {\it end-to-end} learning framework of \citet{factorized_embed} when the architecture is replaced by ours: the only difference when compared to the first row is solely due to the effect of our architecture. In the third row, we present our proposed two-step method. The improvement in performance from the second to the third row is only due to our pre-training method approach: \textit{we emphasize that the neural architectures and landmark losses are exactly the same}. Our approach consistently leads to significantly improved performances. In figure \ref{fig:final_cummulative_error} we plot the final accuracy curves of our pre-training method vs the end-to-end training method. As discussed before, we fit a linear regressor that predicts the location of the ground truth annotations by taking the unsupervised landmark locations as input. The regressor is fitted on a training dataset and then used on a separate test dataset. The accuracy curves in figure \ref{fig:final_cummulative_error} are obtained from the test data predicted vs ground truth manual annotations of the three datasets BBCPose, Cat head, and MAFL.




\section{Conclusions}
Our study shows that the intermediate neural representations learned by standard {\it end-to-end} approach leveraging equivariance losses enjoy poor equivariance properties. 
Instead, for unsupervised landmark discovery, we proposed to use contrastive learning for pre-training equivariant intermediate neural representations.
The numerical experiments demonstrate that this simple strategy, which can naturally be used within several other pipelines and tasks, can lead to a significant performance boost even in challenging situations such as the one described in Section \ref{sec.experiments}.

%
%
\bibliographystyle{plainnat}
\bibliography{egbib}

\end{document}